\documentclass{article}

\usepackage{amsmath}
\usepackage{amssymb}

\usepackage{hyperref}

\usepackage{graphicx} 
\usepackage{epstopdf}
\usepackage{caption}
\usepackage{subcaption}
\usepackage{makecell}

\def\ie{\emph{i.e\@.}\xspace}
\def\eg{\emph{e.g\@.}\xspace}

\usepackage{xcolor}
\usepackage{xspace}
\usepackage{verbatim}
\usepackage{mathtools}
\usepackage{amsthm}
\usepackage{multirow}
\usepackage{tabu}
\usepackage{sidecap}

\usepackage{array}
\newcolumntype{L}[1]{>{\raggedright\let\newline\\\arraybackslash\hspace{0pt}}m{#1}}
\newcolumntype{C}[1]{>{\centering\let\newline\\\arraybackslash\hspace{0pt}}m{#1}}
\newcolumntype{R}[1]{>{\raggedleft\let\newline\\\arraybackslash\hspace{0pt}}m{#1}}

\newtheorem{thrm}{Theorem}

\DeclareMathOperator*{\argmin}{argmin}

\DeclareMathOperator*{\argmax}{argmax}

\newcommand{\aka}{\emph{a.k.a\@.}\xspace}
\newcommand{\sq}[2][0]{
  \mbox{$\medmuskip=#1mu\displaystyle#2$}%
}

\newcolumntype{"}{!{\vrule width 1pt}}

\newcommand{\kmeans}{$k$-means\xspace}
\newcommand{\kmeansbold}{\textbf{$\boldsymbol{k}$-means}\xspace}
\newcommand{\kmeanspp}{$k$-means++\xspace}
\newcommand{\kmeansppbold}{\textbf{$\boldsymbol{k}$-means++}\xspace}
\newcommand{\ourmethod}{G-MM\xspace}
\newcommand{\LSSVM}{Latent Structural SVM\xspace}
\newcommand{\lssvm}{LS-SVM\xspace}
\newcommand{\bound}{b}
\newcommand{\myquote}[1]{``#1''}

\usepackage[accepted]{icml2019}

\icmltitlerunning{Generalized Majorization-Minimization}

\begin{document}

\twocolumn[
\icmltitle{Generalized Majorization-Minimization}



\icmlsetsymbol{equal}{*}

\begin{icmlauthorlist}
\icmlauthor{Sobhan Naderi}{google}
\icmlauthor{Kun He}{frl}
\icmlauthor{Reza Aghajani}{ucsd}
\icmlauthor{Stan Sclaroff}{bu}
\icmlauthor{Pedro Felzenszwalb}{brown}
\end{icmlauthorlist}

\icmlaffiliation{google}{Google Research}
\icmlaffiliation{frl}{Facebook Reality Labs}
\icmlaffiliation{ucsd}{University of California San Diego}
\icmlaffiliation{bu}{Boston University}
\icmlaffiliation{brown}{Brown University}

\icmlcorrespondingauthor{Sobhan Naderi}{sobhan@google.com}


\vskip 0.3in
]



\printAffiliationsAndNotice{* The paper was written when Kun He was at Boston University.} 

\begin{abstract}
Non-convex optimization is ubiquitous in machine learning.
Majorization-Minimization (MM) is a powerful iterative procedure 
for optimizing non-convex functions that works by optimizing 
a sequence of bounds on the function.
In MM, the bound at each iteration is required to \emph{touch} 
the objective function at the optimizer of the previous bound.
We show that this touching constraint is unnecessary and
overly restrictive. We generalize MM by relaxing this constraint,
and propose a new optimization framework, named Generalized
Majorization-Minimization (\ourmethod), that is more flexible. 
For instance, \ourmethod can incorporate application-specific
biases into the optimization procedure without changing the
objective function.
We derive \ourmethod algorithms for several latent variable models
and show empirically that they consistently outperform their MM
counterparts in optimizing non-convex objectives. In particular,
\ourmethod algorithms appear to be less sensitive to initialization.
\end{abstract}

\section{Introduction}
Non-convex optimization is ubiquitous in machine learning.
For example, data clustering  \cite{kmeans, kmeans:plusplus}, training 
classifiers with latent variables \cite{yu09, dpm, Pirsiavash14, glvm},
and training visual object detectors from weakly labeled data 
\cite{Hyun14, Rastegari15, Ries15} all lead to non-convex optimization 
problems.

Majorization-Minimization (MM)~\cite{mm} is an optimization framework 
for designing well-behaved optimization algorithms for non-convex functions. 
MM algorithms work by iteratively optimizing a sequence of easy-to-optimize
surrogate functions that bound the objective.
Two of the most successful instances of MM algorithms are 
Expectation-Maximization (EM) \cite{em} and the Concave-Convex Procedure 
(CCP)~\cite{cccp}. However, both have a number of drawbacks in practice, 
such as sensitivity to initialization and lack of uncertainty modeling for latent 
variables. This has been noted in \cite{gen_em,dpm,naderi12,kumar12,ping14}.

We propose a new procedure,
\emph{Generalized Majorization-Minimization (\ourmethod)},
for non-convex optimization.  Our approach is inspired by MM, but we 
generalize the bound construction process to allow for a \emph{set} of 
valid bounds to be used, while still maintaining algorithmic convergence. 
This generalization gives us more freedom in bound selection and can 
be used to design better optimization algorithms.

In training latent variable models and in clustering problems, MM 
algorithms such as CCP and \kmeans are known to be sensitive to the 
initial values of the latent variables or cluster memberships.  We refer 
to this problem as \emph{stickiness} of the algorithm to the initial latent 
values. Our experimental results show that \ourmethod leads to 
methods that tend to be less sticky to initialization. We demonstrate 
the benefit of using \ourmethod on multiple problems, including 
\kmeans clustering and applications of \LSSVM{s} to image 
classification with latent variables.

\subsection{Related Work}

One of the most popular and well studied iterative methods 
for non-convex optimization is the EM algorithm \cite{em}.
EM is best understood in the context of maximum 
likelihood estimation in the presence of missing data, or 
\textit{latent variables}. EM is a bound optimization algorithm: in 
each E-step, a lower bound on the likelihood is constructed, and 
the M-step maximizes this bound.

Countless efforts have been made to extend the EM algorithm 
since its introduction. In~\cite{gen_em} it is shown that, while 
both steps in EM involve optimizing some functions, it is not 
necessary to fully optimize the functions in each step; in fact, 
each step only needs to \myquote{make progress}. This 
relaxation can potentially avoid sharp local minima and even 
speed up convergence.

The Majorization-Minimization (MM) framework \cite{mm} 
generalizes EM by optimizing a sequence of surrogate functions 
(bounds) on the original objective function.
The Concave-Convex Procedure (CCP)~\cite{cccp} is a 
widely-used instance of MM where the surrogate function is 
obtained by \textit{linearizing} the concave part of the objective.
Many successful learning algorithms employ CCP, \eg the 
Latent SVM~\cite{dpm}. Other instances of MM algorithms 
include iterative scaling~\cite{pietra97}, and non-negative matrix 
factorization~\cite{nmf}. Another related line of research concerns 
the Difference-of-Convex (DC) programming~\cite{dc_prog}, 
which can be shown to reduce to CCP under certain conditions.
Convergence properties of such general \myquote{bound 
optimization} algorithms have been discussed in 
\cite{salakhutdinov02}.

Despite widespread success, MM (and CCP in particular) has a 
number of drawbacks, some of which have motivated our work. 
In practice, CCP often exhibits \emph{stickiness} to initialization, 
which necessitates expensive initialization or multiple trials 
\cite{naderi12, Hyun14, multifold}. In optimizing latent variable 
models, CCP lacks the ability to incorporate application-specific 
information without making modifications to the objective function, 
such as prior knowledge or side information~\cite{xing02,yu12}, 
latent variable uncertainty~\cite{kumar12,ping14}, and posterior 
regularization~\cite{ganchev10}.
Our framework addresses these drawbacks. Our key observation 
is that we can relax the constraint enforced by MM that requires 
the bounds to touch the objective function, and this relaxation 
gives us the ability to better avoid sensitivity to initialization, and 
to incorporate side information.

A closely related work to ours is the \myquote{pseudo-bound}
optimization framework by~\cite{tang14}. It generalizes CCP 
using bounds that may intersect the objective function.
In contrast, our framework uses valid bounds, but only relaxes 
the touching requirement. Also, the pseudo-bound optimization 
framework is specific to binary energies in MRFs (although, it 
was recently generalized to multi-label energies in~\cite{tang18}),
and it restricts the form of surrogate functions to parametric max-flow.

The generalized variants of EM proposed and analyzed by 
\cite{gen_em} and \cite{gam} are related to our work when we 
restrict our attention to probabilistic models and the EM algorithm.
EM can be viewed as a bound optimization procedure where the 
likelihood function involves both the model parameters $\theta$ 
and a distribution $q$ over the latent variables, denoted by 
$F(\theta,q)$. Choosing $q$ to be the posterior leads to a lower 
bound on $F$ that is tight at the current estimate of $\theta$.
Generalized versions of EM, such as those given by~\cite{gen_em},
use distributions other than the posterior in an alternating optimization 
of $F$. This fits into our framework, as we use the exact same 
objective function, and only changes the bound construction step 
(which amounts to picking the distribution $q$ in EM). We propose 
both \emph{stochastic} and \emph{deterministic} strategies for 
bound construction, and demonstrate that they lead to higher 
quality solutions and less sensitivity to initialization than other 
EM-like methods.

\section{Proposed Optimization Framework}
\begin{figure*}
  \centering
  \begin{minipage}[t]{0.48\textwidth}
	\vspace{-0.1cm}
    \centering
    \raisebox{-\height}{\includegraphics[width=0.95\textwidth]{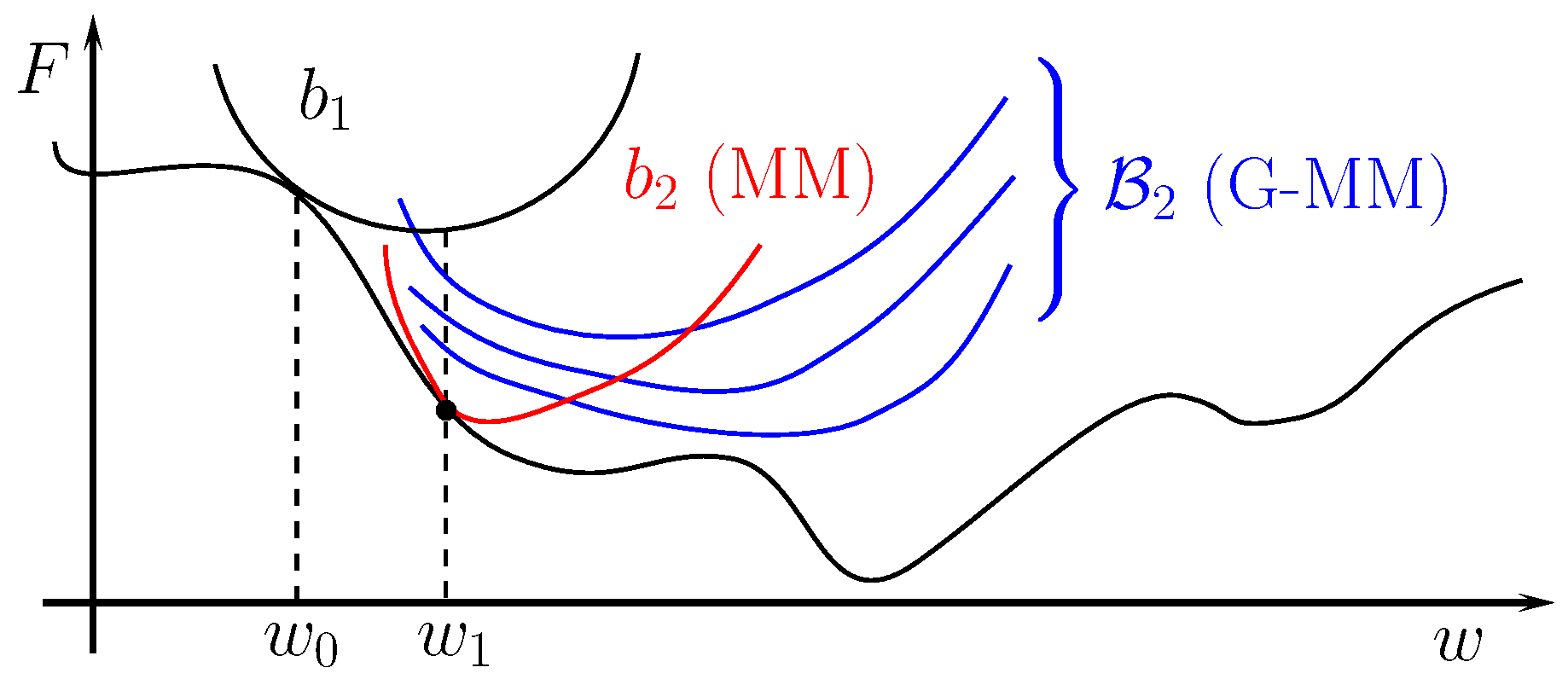}}
    \vspace{-0.3cm}
    \caption{Optimization of $F$ using MM (red) and
      \ourmethod (blue).  In MM the bound $b_2$ has to touch $F$ at
      $w_1$.  In \ourmethod we only require that $b_2$ be below $b_1$
      at $w_1$, leading to several choices $\mathcal{B}_2$.}
	\label{fig:cccp_vs_ours}
  \end{minipage}\hfill
  \begin{minipage}[t]{0.48\textwidth}
    \vspace{-0.3cm}
    \begin{algorithm}[H]
    \small
    \caption{\ourmethod optimization}
    \label{alg:our_framework}
    \begin{algorithmic}[1]
      \INPUT{$w_0, \eta, \epsilon$}
      \STATE $v_0 := F(w_0)$
      \FOR {$t := 1, 2, \dots$}
      \STATE select $\bound_t \in \mathcal{B}_t = \mathcal{B}(w_{t-1}, v_{t-1})$ as in (\ref{eq:valid_bounds_eta})
      \label{alg:our_framework:construct_bound}
      \STATE $w_t := \argmin_w \bound_t(w)$ \label{alg:our_framework:optimize_bound}
      \STATE $d_t := \bound_t(w_t) - F(w_t)$
      \STATE $v_t := \bound_t(w_t) - \eta d_t$ \label{alg:our_framework:vt}
      \STATE \textbf{if} $d_t < \epsilon$ \textbf{break}
      \ENDFOR
      \OUTPUT{$w_t$}
    \end{algorithmic}
    \end{algorithm}
  \end{minipage}
\vspace{-0.3cm}
\end{figure*}

We consider minimization of functions that are bounded from below. The 
extension to maximization is trivial.
Let $\sq{F(w) : \mathbb{R}^d\rightarrow \mathbb{R}}$ be a lower-bounded 
function that we wish to minimize. We propose an iterative procedure that 
generates a sequence of solutions $w_1,w_2,\ldots$ until it converges. The
solution at iteration $\sq{t\geq 1}$ is obtained by minimizing an upper bound 
$\bound_t(w)$ to the objective function \ie $w_t = \argmin_w \bound_t(w)$. 
The bound at iteration $t$ is chosen from a set of \myquote{valid} bounds 
$\mathcal{B}_t$ (see Figure~\ref{fig:cccp_vs_ours}). In practice, we take the 
members of $\mathcal{B}_t$ from a family $\mathcal{F}$ of functions that 
upper-bound $F$ and can be optimized efficiently, such as quadratic functions,
or quadratic functions with linear constraints. $\mathcal{F}$ must be rich 
enough so that $\mathcal{B}_t$ is never empty.
Algorithm~\ref{alg:our_framework} gives the outline of the approach.

This general scheme is used in both MM and \ourmethod. However, as we 
shall see in the rest of this section, MM and \ourmethod have key differences 
in the way they measure progress and the way they construct new bounds.

\subsection{Progress Measure}
MM measures progress with respect to the objective values.
To guarantee progress over time MM requires that the bound
at iteration $t$ must touch the objective function at the previous
solution, leading to the following constraint:
\begin{align}
	\textbf{MM constraint:}
	\hspace{1cm}
	b_t(w_{t-1}) = F(w_{t-1}).
\label{eq:MMconstraint}
\end{align}
This touching constraint, together with the fact that $w_t$ minimizes $b_t$ 
leads to $F(w_t) \le F(w_{t-1})$. That is, the value of the objective function
is non-increasing over time. However, it can make it hard to avoid local minima,
and eliminates the possibility of using bounds that do not touch the objective 
function but may have other desirable properties.

In \ourmethod, we measure progress with respect to the bound values.
It allows us to relax the touching
constraint of MM, stated in (\ref{eq:MMconstraint}), and require instead that,
\begin{align}
\textbf{\ourmethod constraints:}
\hspace{0.3cm}
\begin{cases}
	b_1(w_0) = F(w_0) \\
	b_t(w_{t-1}) \le b_{t-1}(w_{t-1}).
\label{eq:GMMconstraint}
\end{cases}
\end{align}
Note that the \ourmethod constraints are weaker than MM: since $b_{t-1}$ 
is an upper bound on $F$, (\ref{eq:MMconstraint}) implies 
(\ref{eq:GMMconstraint}). While MM constraint implies that the sequence 
$\{F(w_t)\}_t$ is decreasing, \ourmethod only requires $\{b_t(w_t)\}_t$ to 
be decreasing.

\subsection{Bound Construction}
\label{sec:bound_construction}
This section describes line 3 of Algorithm~\ref{alg:our_framework}. To 
construct a bound at iteration $t$, \ourmethod considers a \myquote{valid} 
subset of upper bounds $\mathcal{B}_t \subseteq \mathcal{F}$.
To guarantee convergence, we restrict $\mathcal{B}_t$ to bounds that
are below a threshold $v_{t-1}$ at the previous solution $w_{t-1}$:
\begin{align}
	\mathcal{B}_t &= \mathcal{B}(w_{t-1}, v_{t-1}) \notag \\
	\mathcal{B}(w, v) &= \{ b \in \mathcal{F} \mid b(w) \leq v \}.
	\label{eq:valid_bounds_eta}
\end{align}
Initially, we set $v_0 = F(w_0)$ to ensure that the first bound
\emph{touches} $F$. 
For $t \ge 1$, we set $v_t = \eta F(w_t) + (1 - \eta) b_t(w_t)$ for some 
hyperparameter $\sq{\eta \in (0,1]}$, which we call the 
\emph{progress coefficient}. This guarantees making at least $\eta d_t$ 
progress, where $\sq{d_t = b_t(w_t) - F(w_t)}$ is the gap between the 
bound and the true objective value at $w_t$. Small values of $\eta$ 
allow for gradual exploratory progress while large values of $\eta$ 
greedily select bounds that guarantee immediate progress. When 
$\sq{\eta=1}$ all valid bounds touch $F$ at $w_{t-1}$, corresponding to 
the MM requirement. Note that all the bounds $b \in \mathcal{B}_t$ 
satisfy (\ref{eq:GMMconstraint}).

We consider two scenarios for selecting a bound from $\mathcal{B}_t$. 
In the first scenario we define a bias function
$g:\mathcal{B}_t \times \mathbb{R}^d \rightarrow \mathbb{R}$ that takes
a bound $b \in \mathcal{B}_t$ and a current solution $w \in \mathbb{R}^d$ 
and returns a scalar indicating the goodness of the bound.
We then select the bound with the largest bias value \ie
$\bound_t = \argmax_{\bound \in \mathcal{B}_t} g(\bound, w_{t-1})$.
In the second scenario we propose to choose a bound from 
$\mathcal{B}_t$ at random. Thus, we have both a deterministic 
(the $1^{st}$ scenario) and a stochastic (the $2^{nd}$ scenario) bound
construction mechanism.

\subsection{Generalization over MM}
MM algorithms, such as EM and CCP, are special cases of 
\ourmethod that use a specific bias function $\sq{g(b,w) = -b(w)}$.
Note that $b_t = \argmax_{b \in \mathcal{B}_t} -b(w_{t-1})$ touches 
$F$ at $w_{t-1}$, assuming $\mathcal{B}_t$ includes such a bound.
Also, by definition, $b_t$ makes maximum progress with respect to 
the previous bound value $b_{t-1}(w_{t-1})$. By choosing bounds 
that maximize progress, MM algorithms tend to rapidly converge to 
a nearby local minimum.
For instance, at iteration $t$, the CCP bound for latent SVMs is 
obtained by fixing the latent variables in the concave part of $F$ 
to the maximizers of the score of the model from iteration $\sq{t-1}$,
making $w_t$ attracted to $w_{t-1}$.
Similarly, in the E-step of EM, the posterior distribution of the 
latent variables is computed with respect to $w_{t-1}$ and, in the 
M-step, the model is updated to \myquote{match} these fixed 
posteriors. This explains one reason why MM algorithms are 
observed to be sticky to initialization.

\ourmethod offers a more flexible bound construction scheme than MM.
In Section~\ref{sec:experiments} we show empirically that picking a valid 
bound randomly, \ie $\sq{b_t \sim U(\mathcal{B}_t)}$, is less sensitive 
to initialization and leads to better results compared to CCP and EM.
We also show that using good bias functions can further improve 
performance of the learned models.

\section{Convergence of \ourmethod}
We show that, under general assumptions, the sequence $\{w_t\}_t$ of 
bound minimizers converges, and Algorithm~\ref{alg:our_framework} 
stops after finite steps (Theorem~\ref{thrm:gmm_sol_converges}). With
additional assumptions, we also prove that the limit of this sequence is 
a stationary point of $F$ 
(Theorem~\ref{thrm:GMM-converges-to-local-min}).
We believe stronger convergence properties depend on the structure 
of the function $F$, the family of the bounds $\mathcal{F}$, and the 
bound selection strategy, and should be investigated separately for 
each specific problem. We prove Theorems 
\ref{thrm:gmm_sol_converges} and 
\ref{thrm:GMM-converges-to-local-min} in the supplementary material.

\begin{thrm}
	Suppose $F$ is a lower-bounded, continuous function with compact 
	sublevel sets, and $\mathcal{F}$ is a family of lower-bounded and 
	$m$-strongly convex functions. Then the sequence of minimizers 
	$\{w_t\}_t$ converges (\ie the limit 
	$w^\dagger = \lim_{t \rightarrow \infty} w_t$ exists), and the gap 
	$\sq{d_t = b_t(w_t) - F(w_t)}$ converges to 0.
	\label{thrm:gmm_sol_converges}
\end{thrm}

\begin{thrm}
	Suppose the assumptions in Theorem~\ref{thrm:gmm_sol_converges} holds.
	In addition, let $F$ be continuously differentiable,
	and $\mathcal{F}$ be a family of smooth functions such that
	$\forall b \in \mathcal{F}, M I \succeq \nabla^2 b(w) \succeq m I$, for some 
	$m, M \in (0, \infty)$, 
	where $I$ is the identity matrix. Then $\nabla F(w^\dagger) = 0$,
	namely,  \ourmethod converges to a stationary point of $F$.
	\label{thrm:GMM-converges-to-local-min}
\end{thrm}

\section{Derived Optimization Algorithms}
\label{sec:derived_algorithms}
\ourmethod is applicable to a variety of non-convex optimization problems,
but for simplicity and ease of exposition, we primarily focus on 
\emph{latent variable models} where bound construction
naturally corresponds to {imputing} latent variables in the model.
In this section we derive \ourmethod algorithms for two widely used
families of models, namely, \kmeans and \LSSVM.
Note that the training objectives of these two problems are non-differentiable
and, therefore, Theorem~\ref{thrm:GMM-converges-to-local-min} does not apply
to them.
However, note that the theorem only gives a \emph{sufficient} condition but is
not necessary for convergence of the algorithms. In fact, in all our experiments
we observe that \ourmethod converges (\eg see Table~\ref{table:CCPvsGMM_cntIters}),
and these algorithms significantly outperform their MM counterparts 
(see Section~\ref{sec:experiments}).

\subsection{\kmeansbold Clustering}
Let $\sq{\{x_1, \dots, x_n\}}$ denote $n$ points and $\sq{w=(\mu_1, \dots,
\mu_k)}$ denote $k$ cluster centers. We assign a cluster to each
point, denoted by $\sq{z_i \in \{1, \dots, k\}}, \forall i \in \{1, \dots, n\}$.
The objective function in \kmeans clustering is defined as follows,
\begin{align}
	F(w) = \sum_{i=1}^n \min_{z_i \in \{1, \dots, k\}} || x_i - \mu_{z_i} ||^2,
	\quad w=(\mu_1, \dots, \mu_k).
\label{eq:kmeans_objective}
\end{align}
\textbf{Bound construction:} We obtain a convex upper bound on $F$ by fixing
the latent variables $(z_1, \dots, z_n)$ to certain values instead of minimizing
over these variables. Such bounds are quadratic convex functions of $(\mu_1,\ldots,\mu_k)$,
\begin{align}
	\mathcal{F} = \left\{\sum_{i=1}^n || x_i - \mu_{z_i} ||^2 \;\middle|\; \forall i,\; z_i \in \{1,\ldots, k\} \right\}.
	\label{eq:boundFamily_kmeans}
\end{align}
The \kmeans algorithm is an instance of MM methods.
The algorithm repeatedly assigns each example to its nearest center
to construct a bound, and then updates the centers by optimizing the
bound. We can set $g(\bound, w) = -\bound(w)$ in \ourmethod to
obtain the \kmeans algorithm. We can also define $g$ differently
to obtain a \ourmethod algorithm that exhibits other desired
properties. For instance, a common issue in clustering is cluster
starvation.  One can discourage starvation by defining $g$ accordingly.

We select a bound from $\mathcal{B}_t$ uniformly at random by
starting from an initial configuration $\sq{z=(z_1, \dots, z_n)}$ that 
corresponds to a valid bound in $\mathcal{B}_t$ (\eg \kmeans solution). 
We then do a random walk on a graph whose nodes are latent
configurations defining valid bounds.  The neighbors of a latent
configuration $z$ are other latent configurations that can be obtained
by changing the value of one of the $n$ latent variables in $z$.

\textbf{Bound optimization:} Optimization of $\bound \in
\mathcal{F}$ can be done in closed form by setting $\mu_j$ to be
the mean of all examples assigned to cluster $j$:
\begin{align}
	\mu_j = \frac{\sum_{i \in I_j} x_i}{|I_j|}, \quad I_j =\{1 \leq i \leq n \mid z_i = j\}.
	\label{eq:kmeans_update}
\end{align}

\subsection{\LSSVM}
A \LSSVM (\lssvm)~\cite{yu09} defines a structured output classifier
with latent variables. It extends the Structural SVM~\cite{joachims09}
by introducing latent variables.

Let $\{ (x_1, y_1), \dots, (x_n, y_n) \}$ denote a set of labeled
examples with $x_i \in \mathcal{X}$ and $y_i \in \mathcal{Y}$.
We assume that each example $x_i$ has an associated latent
value $z_i \in \mathcal{Z}$. Let $\phi(x, y, z):\mathcal{X} \times
\mathcal{Y} \times \mathcal{Z} \rightarrow \mathbb{R}^d$ denote
a \emph{feature map}.  A vector $w \in \mathbb{R}^d$ defines
a classifier $\hat{y}:\mathcal{X} \rightarrow \mathcal{Y}$,
\begin{align}
  \hat{y}(x) = \argmax_y ( \max_z w \cdot \phi(x,y,z) ).
\end{align}
The LS-SVM training objective is defined as follows,
\begin{align}
	F(w) = \frac{\lambda}{2} ||w||^2 + \frac{1}{n}\sum_{i=1}^n \hspace{-0.05cm} \Big( \hspace{-0.05cm} & \max_{y, z} \left( w \cdot \phi(x_i, y, z) + \Delta(y, y_i) \right) \notag \\
	& - \max_z w \cdot \phi(x_i, y_i, z) \Big),
	\label{eq:LSSVM_objective}
\end{align}
where  $\lambda$ is a hyper-parameter that controls regularization
and $\Delta(y, y_i)$ is a non-negative loss function that penalizes
the prediction $y$ when the ground truth label is $y_i$.

\textbf{Bound construction:} As in the case of \kmeans,
a convex upper bound on the LS-SVM objective can be obtained
by imputing latent variables. Specifically, for each example $x_i$,
we fix $z_i \in \mathcal{Z}$, and replace the maximization in the
last term of the objective with a linear function $w \cdot \phi(x_i, y_i, z_i)$.
This forms a family of convex piecewise quadratic bounds,
\begin{align}
	\hspace{-0.23cm}\mathcal{F} \hspace{-0.1cm} = \hspace{-0.1cm} \Bigg\{ \hspace{-0.1cm} \frac{\lambda}{2} ||w||^2 + \frac{1}{n} \hspace{-0.05cm} \sum_{i=1}^n & \max_{y, z}\left( w \cdot \phi(x_i, y, z) + \Delta(y, y_i) \right) \notag \\
	& - w \cdot \phi(x_i, y_i, z_i) \Bigg| \forall i, z_i \in \mathcal{Z} \hspace{-0.05cm} \Bigg\}.
	\label{eq:boundFamily_lssvm}
\end{align}
The CCP algorithm for \lssvm selects the bound $b_t$ defined
by $z^t_i=\argmax_{z_i} w_{t-1} \cdot \phi(x_i, y_i, z_i)$. This
particular choice is a special case of \ourmethod when
$g(\bound, w) = -\bound(w)$.

To generate random bounds from $\mathcal{B}_t$
we use the same approach as in the case of \kmeans clustering.
We perform a random walk in a graph where the nodes are latent
configurations leading to valid bounds, and the edges connect
latent configurations that differ in a single latent variable.

\textbf{Bound optimization:} Optimization of $\bound \in \mathcal{F}$ 
corresponds to a convex quadratic program and can be solved using 
different techniques, including gradient based methods (\eg SGD) 
and the cutting-plane method~\cite{joachims09}. 
We use the cutting-plane method in our experiments.

\subsection{Bias Function for Multi-fold MIL}
\label{sec:bias_for_mfmil}
The multi-fold MIL algorithm \cite{multifold} was introduced for
training latent SVMs for weakly supervised object localization,
to deal with stickiness issues in training with CCP. It modifies
how latent variables are updated during training.  \cite{multifold}
divide the training set into $K$ folds, and updates the latent
variables in each fold using a model trained on the other $K-1$
folds. This algorithm does not have a formal convergence guarantee.
By defining a suitable bias function, we can derive a \ourmethod
algorithm that mimics the behavior of multi-fold MIL, and yet, is
convergent.

Consider training an \lssvm. Let $\sq{S=\{1,\ldots,n\}}$ and $I \subseteq S$
denote a subset of $S$.
Also, let $\sq{z_i \in \mathcal{Z}}$ denote the latent
variable associated to training example $(x_i, y_i)$, and $z_{I}^t$ denote
the fixed latent values for training examples indexed by $I$ in iteration $t$.
We denote the model trained on $\{(x_i,y_i) | i\in I\}$ with latent variables
fixed to $z_{I}^t$ in the last maximization of (\ref{eq:LSSVM_objective}) by
$w(I, z_{I}^t)$.

We assume access to a loss function $\ell(w,x,y,z)$. For example,
for the binary latent SVM where $y\in\{-1,1\}$, $\ell$ is the hinge
loss: $\ell(w,x,y,z) = \max\{0,1-y\, w\cdot\phi(x,z)\}$.

We first consider the Leave-One-Out (LOO) setting, \ie $K=n$, and call the algorithm of \cite{multifold} LOO-MIL in this case.
The update rule of LOO-MIL in iteration $t$ is to set
\begin{align}
z_i^t=\argmin_{z\in\mathcal{Z}}~\ell\left(w(S\backslash i,z_{S\backslash i}^{t-1}),x_i,y_i,z\right), ~~\forall i\in S.
\label{eq:update_rule}
\end{align}
After updating the latent values for all training examples, the
model $w$ is retrained by optimizing the resulting bound.

Now let us construct a \ourmethod bias function that mimics
the behavior of LOO-MIL. Recall from
(\ref{eq:boundFamily_lssvm}) that each bound $\sq{b \in \mathcal{B}_t}$
is associated with a joint latent configuration $\sq{z(b)=(z_1,\ldots,z_n)}$.
We use the following bias function:
\begin{align}
g(b,w) = -\sum_{i\in S} \ell\left(w(S\backslash i,z_{S\backslash i}^{t-1}),x_i,y_i,z_i\right).
\label{eq:loomil}
\end{align}
Note that picking a bound according to (\ref{eq:loomil}) is
equivalent to the LOO-MIL update rule of (\ref{eq:update_rule})
except that in (\ref{eq:loomil}) only \emph{valid} bounds are
considered; that is bounds that make at least $\eta$-progress.

For the general multi-fold case (\ie $K<n$), the bias function can
be derived similarly.


\section{Experiments}
\label{sec:experiments}

We evaluate \ourmethod and MM algorithms on \kmeans clustering and \lssvm training on
various datasets. Recall from (\ref{eq:valid_bounds_eta}) that the progress
coefficient $\eta$ defines the set of valid bounds $\mathcal{B}_t$ in each step.
CCP and standard \kmeans bounds correspond to setting $\sq{\eta = 1}$, thus taking
maximally large steps towards a local minimum of the true objective.


\subsection{\kmeansbold Clustering}
\label{sec:experiments:clustering}

We conduct experiments on four clustering datasets: Norm-25~\cite{kmeans:plusplus},
D31~\cite{kmeans:D31}, Cloud~\cite{kmeans:plusplus}, and GMM-200.
See the references for details about the datasets. GMM-200 was created by us 
and has $10000$ samples taken from a $2$-D Gaussian mixture model with 200 
mixtures (50 samples per each component). All the mixture components have unit 
variance and their means are placed on a $\sq{70 \times 70}$ square uniformly 
at random, while making sure the distance between any two centers is at least 
$2.5$.

\begin{table*}[t]
\small
\centering
	\renewcommand{\tabcolsep}{3pt}
	\begin{tabular}{| c | c | c | c   | c |    c | c | c | c | }
	\hline
	\multirow{2}{*}{\textbf{dataset}} & \multirow{2}{*}{\textbf{k}} & \textbf{opt.} &
	\multicolumn{2}{c|}{\textbf{forgy}} & \multicolumn{2}{c|}{\textbf{random partition}} & \multicolumn{2}{c|}{\kmeansppbold} \\ \cline{4-9}
	& & \textbf{method} & \textbf{avg $\pm$ std} & \textbf{best} & \textbf{avg $\pm$ std} & \textbf{best} & \textbf{avg $\pm$ std} & \textbf{best} \\ \hline
	\multirow{2}{*}{Norm-25} & \multirow{2}{*}{$25$} & \kmeans
	& $\sq{1.9e5 \pm 2e5}$ & $\sq{7.0e4}$
	& $\sq{5.8e5 \pm 3e5}$ & $\sq{2.2e5}$
	& $\sq{5.3e3 \pm 9e3}$ & $\sq{1.5}$  \\ \cline{3-9}
	& & \ourmethod
	& $\sq{9.7e3 \pm 1e4}$ & $\sq{1.5}$
	& $\sq{2.0e4 \pm 0}$ & $\sq{2.0e4}$
	& $\sq{4.5e3 \pm 8e3}$ & $\sq{1.5}$ \\ \hline
	\multirow{2}{*}{D31} & \multirow{2}{*}{$31$} & \kmeans
	& $1.69 \pm 0.03$ & $1.21$
	& $52.61 \pm 47.06$ & $4.00$
	& $1.55 \pm 0.17$ & ~ $1.10$ ~ \\ \cline{3-9}
	& & \ourmethod
	& $1.43 \pm 0.15$ & $1.10$
	& $1.21 \pm 0.05$ & $1.10$
	& $1.45 \pm 0.14$ & $1.10$ \\ \hline
	\multirow{2}{*}{Cloud} & \multirow{2}{*}{$50$} & \kmeans
	& $1929 \pm 429$ & $1293$
	& $44453 \pm 88341$ & $3026$
	& $1237 \pm 92$ & $1117$ \\ \cline{3-9}
	& & \ourmethod
	& $1465 \pm 43$ & $1246$
	& $1470 \pm 8$ & $1444$
	& $1162 \pm 95$ & $1067$ \\ \hline
	\multirow{2}{*}{ GMM-200 } & \multirow{2}{*}{ $200$ } & \kmeans
	& $2.25 \pm 0.10$ & $2.07$
	& $11.20 \pm 0.63$ & $9.77$
	& $2.12 \pm 0.07$ & $1.99$ \\ \cline{3-9}
	& & \ourmethod
	& $2.04 \pm 0.09$ & $1.90$
	& $1.85 \pm 0.02$ & $1.80$
	& $1.98 \pm 0.06$ & $1.89$ \\ \hline
	\end{tabular}
\vspace{-0.7em}
\caption{Comparison of \ourmethod and \kmeans on four clustering datasets
	  and three initialization methods; \textbf{forgy} initializes cluster centers to
	  random examples, \textbf{random partition} assigns each data point to a random
	  cluster center, and \kmeansppbold implements the algorithm from~\cite{kmeans:plusplus}.
	  The mean, standard deviation, and best objective values out of $50$ random trials
	  are reported. \kmeans and \ourmethod use the exact same initialization in
	  each trial. \ourmethod consistently converges to better solutions.}
\label{table:kmeans}
\end{table*}

We compare results from three different initializations: \textbf{forgy} selects
$k$ training examples uniformly at random without replacement to define initial cluster centers, \textbf{random partition} assigns
training samples to cluster centers randomly, and \kmeansppbold uses the algorithm in~\cite{kmeans:plusplus}.
In each experiment we run the algorithm 50 times and report the mean, standard deviation, and the best objective
value (\ref{eq:kmeans_objective}). Table~\ref{table:kmeans} shows the results using \kmeans
(hard-EM) and \ourmethod.
We note that the variance of the solutions found by \ourmethod is typically smaller than \kmeans. Moreover, the best and the
average solutions found by \ourmethod are always better than (or the same as) those found by \kmeans.
This trend generalizes over different initialization schemes as well as different datasets.

Although \emph{random partition} seems to be a bad initialization for \kmeans on all datasets, \ourmethod
recovers from it. In fact, on D31 and GMM-200 datasets, \ourmethod initialized by \emph{random partition}
performs better than when it is initialized by other methods (including \kmeanspp). Also, the
variance of the best solutions (across different initialization methods) in \ourmethod is smaller than that of \kmeans.
These suggest that the \ourmethod optimization is less sticky to initialization than \kmeans.

Figure~\ref{fig:different_progress_coeffs} shows the effect of the progress coefficient on the quality of the solution found
by \ourmethod. Different initialization schemes are color coded. The solid line indicates the average objective
over 50 iterations, the shaded area covers one standard deviation from the average, and the dashed line indicates the
best solution over the 50 trials. Smaller progress coefficients allow for more extensive exploration, and hence, smaller
variance in the quality of the solutions. On the other hand, when the progress coefficient is large \ourmethod is more
sensitive to initialization (\ie is more sticky) and, thus, the quality of the solutions over multiple runs is more
diverse. However, despite the greater diversity, the best solution is worse when the progress coefficient is large.
\ourmethod reduces to \kmeans if we set the progress coefficient to 1 (\ie the largest possible value).

\iffalse
\begin{figure*}[t]
\begin{subfigure}[b]{0.4\textwidth}
\centering
\includegraphics[height=1.75in]{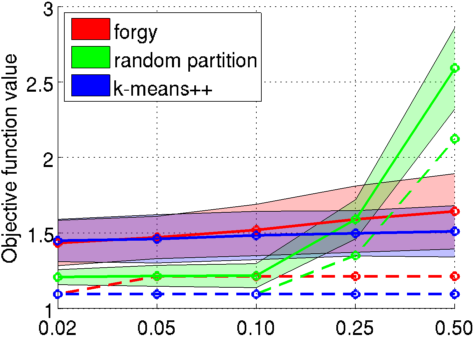}
\caption{D31}
\end{subfigure}
\hfill
\begin{subfigure}[b]{0.4\textwidth}
\centering
\includegraphics[height=1.75in]{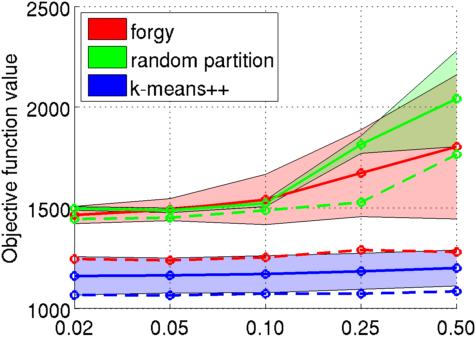}
\caption{Cloud}
\end{subfigure}
	\caption{Effect of the progress coefficient $\eta$ (x-axis) on the quality of
	the solutions found by \ourmethod (y-axis) on two clustering
	datasets. The quality is measured by the objective function in
	(\ref{eq:kmeans_objective}). Lower values are better. The average (solid
	line), the best (dashed line), and the variance (shaded area) over 50 trials
	are shown in the plots and different initializations are color coded.}
	\label{fig:different_progress_coeffs}
\end{figure*}
\else
\begin{SCfigure*}[1.0][t]
\centering
	\begin{subfigure}[b]{0.36\textwidth}
		\centering
		\includegraphics[width=2.5in]{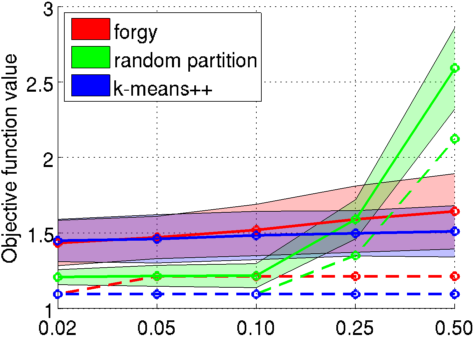}
		\caption{D31}
	\end{subfigure}
	\begin{subfigure}[b]{0.36\textwidth}
		\centering
		\includegraphics[width=2.5in]{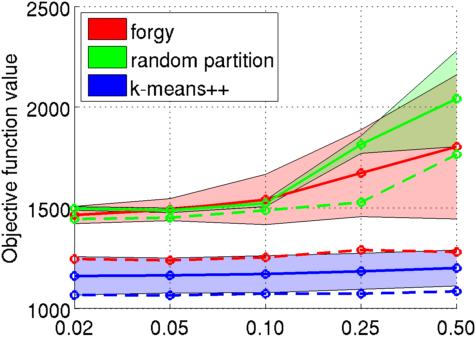}
		\caption{Cloud}
	\end{subfigure}
	\caption{Effect of the progress coefficient $\eta$ (x-axis) on the quality of
	the solutions found by \ourmethod (y-axis) on two clustering
	datasets. The quality is measured by the objective function in
	(\ref{eq:kmeans_objective}). Lower values are better. The average (solid
	line), the best (dashed line), and the variance (shaded area) over $50$ trials
	are shown in the plots and different initializations are color coded.}
	\label{fig:different_progress_coeffs}
\vspace{-1.2em}
\end{SCfigure*}
\fi

\subsection{\LSSVM for Image Classification and Object Detection}
We consider the problem of training an \lssvm classifier on the mammals dataset ~\cite{mammals}.
The dataset contains images of six mammal categories with image-level annotation. Locations of the objects
in these images are not provided, and therefore, treated as latent variables in the model. Specifically,
let $x$ be an image and $y$ be a class label ($y \in \{1,\ldots,6\}$ in this case), and let $z$ be the latent
location of the object in the image. 
We define $\phi(x, y, z)$ to be a feature function with $6$ blocks;
one block for each category. It extracts features from location $z$ of image $x$ and places them in the
$y$-th block of the output and fills the rest with zero. We use the following multi-class classification rule:
\begin{align}
	y(x) = \argmax_{y, z} w \cdot \phi(x, y, z), \; w = (w_1, \ldots, w_6).
\end{align}
In this experiment we use a setup similar to that in~\cite{kumar12}: we use Histogram of Oriented Gradients
(HOG) for the image feature $\phi$, and the 0-1 classification loss for $\Delta$. We set
$\lambda=0.4$ in (\ref{eq:LSSVM_objective}). We report 5-fold
cross-validation performance. Three initialization strategies are considered for the latent object locations:
\emph{image center}, \emph{top-left corner}, and \emph{random locations}. The first is a reasonable initialization
since most objects are at the center in this dataset; the second initialization strategy is somewhat adversarial.

We try a stochastic as well as a deterministic bound construction
method.  For the stochastic method, in each iteration $t$ we
uniformly sample a subset of examples $S_t$ from the training
set, and update their latent variables using $z^t_i = \argmax_{z_i}
w_{t-1} \cdot \phi(x_i, y_i, z_i)$. Other latent variables are kept
the same as the previous iteration. We increase the size of $S_t$
across iterations.

For the deterministic method, we use the bias function that we
described in Section~\ref{sec:bias_for_mfmil}. This is inspired
by the multi-fold MIL idea \cite{multifold} and is shown to
reduce stickiness to initialization, especially in high dimensions.
We set the number of folds to $\sq{K=10}$ in our experiments.

\begin{table*}[t]
\centering
\small
\renewcommand{\tabcolsep}{4pt}
\begin{tabular}{|l|c|c|c|c|c|c|c|c|c|c|}
\hline
\multirow{2}{*}{\textbf{Opt. Method}} & \multicolumn{2}{c|}{\textbf{center}} & \multicolumn{2}{c|}{\textbf{top-left}} & \multicolumn{2}{c|}{\bf random} \\ \cline{2-7}
 & \textbf{objective} &  \textbf{test error} & \textbf{objective} &  \textbf{test error} & \textbf{objective} &  \textbf{test error} \\ \hline
\textbf{CCP}						& 1.21 $\pm$ 0.03 & 22.9 $\pm$ 9.7 & 1.35 $\pm$ 0.03 & 42.5 $\pm$  \;\;4.6 & 1.47 $\pm$ 0.03 & 31.8 $\pm$ 2.6 \\ \hline
\textbf{\ourmethod random}	& 0.79 $\pm$ 0.03 & 17.5 $\pm$ 3.9 & 0.91 $\pm$ 0.02 & 31.4 $\pm$      10.1 & 0.85 $\pm$ 0.03 & 19.6 $\pm$ 9.2 \\ \hline
\textbf{\ourmethod biased}		& 0.64 $\pm$ 0.02 & 16.8 $\pm$ 3.2 & 0.70 $\pm$ 0.02 & 18.9 $\pm$  \;\;5.0 & 0.65 $\pm$ 0.02 & 14.6 $\pm$ 5.4 \\ \hline
\end{tabular}
\vspace{-0.7em}
\caption{\lssvm results on the mammals dataset~\cite{mammals}. We report the mean and standard deviation
			of the training objective (\ref{eq:LSSVM_objective}) and test error over five folds. Three strategies
			for initializing latent object locations are tried: image center, top-left corner, and random location.
			\myquote{\ourmethod random} uses random bounds, and \myquote{\ourmethod bias} uses a bias function inspired by multi-fold MIL \cite{multifold}. Both variants consistently and significantly outperform
			the CCP baseline.}
\label{table:mammal}
\end{table*}

\begin{figure*}[t]
\includegraphics[width=0.33\linewidth]{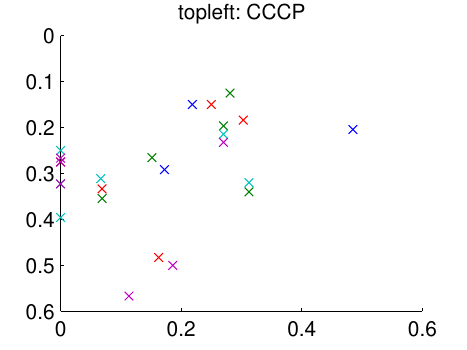}
\hfill
\includegraphics[width=0.33\linewidth]{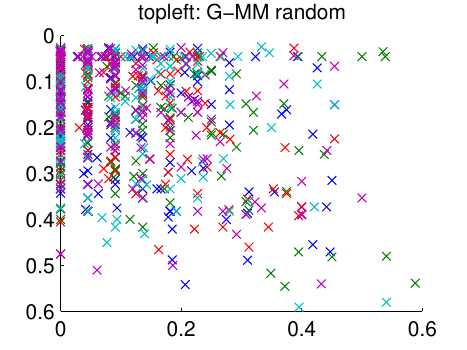}
\hfill
\includegraphics[width=0.33\linewidth]{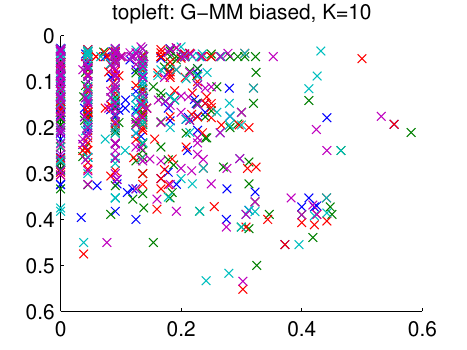}
\vspace{-2.5em}
\caption{Latent location changes  after learning, in relative 
image coordinates, for all five cross-validation folds, for the 
\emph{top-left} initialization on the mammals dataset. Left to 
right: CCP, \myquote{\ourmethod random}, \myquote{\ourmethod 
biased} ($K$=10). Each cross represents a training image; 
cross-validation folds are color coded differently. Averaged over 
five folds, CCP only alters $2.4\%$ of all latent locations, leading 
to very bad performance. \myquote{\ourmethod random} and 
\myquote{\ourmethod biased} alter $86.2\%$ and $93.6\%$ on 
average, respectively, and perform much better.}
\label{fig:latent_locations}
\vspace{-1em}
\end{figure*}

Table~\ref{table:mammal} shows results on the mammals dataset.
Both variants of \ourmethod consistently outperform CCP in terms
of training objective and test error. We observed that CCP  rarely
updates  the latent locations, under all initializations. On the other
hand, both variants of \ourmethod significantly alter the latent
locations, thereby avoiding the local minima close
to the initialization. Figure~\ref{fig:latent_locations} visualizes this
for \emph{top-left} initialization. Since objects rarely occur at the
top-left corner in the mammals dataset, a good model is expected
to significantly update the latent locations.
Averaged over five cross-validation folds, about $90\%$ of the latent
variables were updated in \ourmethod after training whereas this
measure was $2.4\%$ for CCP. This is consistent with the better
training objectives and test errors of \ourmethod.

\subsection{\LSSVM for Scene Recognition}
\label{sec:experiments:scene_classification}
We implement the reconfigurable model of~\cite{naderi12} 
(called RBoW) to do scene classification on MIT-Indoor 
dataset~\cite{mit_indoor}, which has images from 67 indoor 
scene categories. We segment each image into a 
$\sq{10 \times 10}$ regular grid and treat the grid cells as 
image regions. We train a model with 200 shared parts. Any 
part can be used to describe the data in a region. We use 
the activations of the 4096 neurons at the penultimate layer 
of the pre-trained hybrid ConvNet of~\cite{places} to extract 
features from image regions and use PCA to reduce the 
dimensionality of the features to 240.

The RBoW model is an instance of \lssvm models. The latent 
variables are the assignments of parts to image regions and 
the output structure is the multi-valued category label 
predictions. \lssvm{s} are known to be sensitive to initialization 
(\aka the stickiness issue). To cope with this issue~\cite{naderi12} 
uses a generative version of the model to initialize the training 
of the discriminative model. Generative models are typically less 
sticky but perform worse in practice. To validate the hypothesis 
regarding stickiness of \lssvm{s} we train models with several 
initialization strategies.

Initializing training entails the assignment of parts to image 
regions \ie setting $z_i$'s in (\ref{eq:boundFamily_lssvm}) to 
define the first bound. To this end we first discover 200 parts 
that capture discriminative features in the training data. We 
then run graph cut on each training image to obtain part 
assignments to image regions.
Each cell in the $\sq{10 \times 10}$ image grid is a node in 
the graph. Two nodes in the graph are connected if their 
corresponding cells in the image grid are next to each other. 
Unary terms in the graph cut are the dot product scores 
between the feature vector extracted from an image region 
and a part filter plus the corresponding region-to-part 
assignment score. Pairwise terms in the graph cut implement 
a \emph{Potts} model that encourages coherent labelings. 
Specifically, the penalty of labeling two neighboring nodes 
differently is $\lambda$ and it is zero otherwise. $\lambda$ 
controls the coherency of the initial assignments. 
We experiment using $\lambda \in \{ 0, 0.25, 0.5, 1 \}$. We 
also experiment with random initialization, which corresponds 
to assigning $z_i$'s randomly. This is the simplest form of 
initialization and does not require discovering initial part 
filters.

\begin{table*}[t]
\small
\centering
	\renewcommand{\tabcolsep}{4pt}
	\begin{tabular}{| l " c|c " c|c " c|c " c|c " c|c |}
	\hline
	\multirow{2}{*}{\textbf{Opt. Method}} & \multicolumn{2}{c"}{\textbf{Random}} & \multicolumn{2}{c"}{\textbf{$\boldsymbol{\lambda=0.00}$}} & \multicolumn{2}{c"}{\textbf{$\boldsymbol{\lambda=0.25}$}} & \multicolumn{2}{c"}{\textbf{$\boldsymbol{\lambda=0.50}$}} & \multicolumn{2}{c|}{\textbf{$\boldsymbol{\lambda=1.00}$}} \\ \cline{2-11}
	 & \textbf{Acc.\% $\pm$ std} & \textbf{O.F.} & \textbf{Acc. \%} & \textbf{O.F.} & \textbf{Acc. \%} & \textbf{O.F.} & \textbf{Acc. \%} & \textbf{O.F.} & \textbf{Acc. \%} & \textbf{O.F.}  \\ \Xhline{2\arrayrulewidth}
	\textbf{CCP} & $41.94 \pm 1.1$ & 15.20 & $40.88$ & 14.81 & $43.99$ & 14.77 & $45.60$ & 14.72 & $46.62$ & 14.70 \\ \hline
	\textbf{\ourmethod random} & $47.51 \pm 0.7$ & 14.89 & $43.38$ & 14.71 & $44.41$ &14.70  & $47.12$ & 14.66 & $49.88$ & 14.58 \\ \hline
	\textbf{\ourmethod biased}  & $49.34 \pm 0.9$  & 14.55 & $44.83$ & 14.63 & $48.07$ & 14.51 & $53.68$ & 14.33 & $56.03$ & 14.32 \\ \hline
	\end{tabular}
	\vspace{-0.7em}
	\caption{Performance of \lssvm trained with CCP and \ourmethod
	on MIT-Indoor dataset. We report classification accuracy (Acc.\%)
	and the training objective value (O.F.). Columns correspond to
	different initialization schemes. \myquote{Random} assigns random
	parts to regions. ${\lambda}$ controls the coherency of the initial
	part assignments: $\sq{\lambda=1}$ ($\sq{\lambda=0}$) corresponds
	to the most (the least) coherent case. \myquote{\ourmethod random}
	uses random bounds and \myquote{\ourmethod biased} uses the bias
	function of (\ref{eq:bias_function:LSSVM}). $\eta=0.1$ in all
	the experiments. Coherent initializations lead to better models in general,
	but, they require discovering good initial parts. \myquote{\ourmethod}
	outperforms CCP, especially with random initialization.
	\myquote{\ourmethod biased} performs the best.
  }
	\label{table:lssvm}
\vspace{-0.7em}
\end{table*}

We do \ourmethod optimization using both random and biased 
bounds. For the latter we use a bias function $g(\bound, w)$ 
that measures coherence of the labeling from which the bound 
was constructed. Recall from (\ref{eq:boundFamily_lssvm}) that 
each bound in $\bound \in B_t$ corresponds to a labeling of the 
image regions. We denote the labeling corresponding to the 
bound $\bound$ by $z(\bound)=(z_1, \dots, z_n)$ where 
$z_i=(z_{i, 1}, \dots, z_{i, 100})$ specifies part assignments for 
all the 100 regions in the $i$-th image. Also, let $E(z_i)$ denote 
a function that measures coherence of the labeling $z_i$. 
In fact, $E(z_i)$ is the Potts energy function on a graph whose 
nodes are $z_{i, 1}, \dots, z_{i, 100}$. The graph respects a 
4-connected neighborhood system (recall that $z_{i, r}$ 
corresponds to the $r$-th cell in the $\sq{10 \times 10}$ grid 
defined on the $i$-th image). If two neighboring nodes $z_{i, r}$ 
and $z_{i, s}$ get different labels the energy $E(z_i)$ increases 
by 1. For biased bounds we use the following bias function 
which favors bounds that correspond to more coherent labelings:
\begin{equation}
	g(\bound, w) = - \sum_{i=1}^n E(z_i), \hspace{0.5cm} z(\bound) = (z_1, \dots, z_n).
	\label{eq:bias_function:LSSVM}
\end{equation}

Table~\ref{table:lssvm} compares performance of models 
trained using CCP and \ourmethod with random and biased 
bounds. For \ourmethod with random bounds we repeat the 
experiment five times and report the average over these five 
trials. Also, for random initialization, we do five trials using 
different random seeds and report the mean and standard 
deviation of the results. \ourmethod does better than CCP 
under \emph{all} initializations. It also converges to a 
solution with lower training objective value than CCP.
Our results show that picking bounds uniformly at random 
from the set of valid bounds is slightly (but consistently) 
better than committing to the CCP bound. We get a 
remarkable boost in performance when we use a 
reasonable prior over bounds (\ie the bias function of 
(\ref{eq:bias_function:LSSVM})). With $\lambda = 1$, CCP 
attains accuracy of  46.6\%, whereas G-MM attains 49.9\%, 
and 56.0\% accuracy with random and biased initialization 
respectively. Moreover, \ourmethod is less sensitive to 
initialization.

\subsection{Running Time}
\ourmethod bounds make a fraction of the progress that can 
be made in each bound construction step. Therefore, we would 
expect \ourmethod to require more steps to converge when 
compared to MM. We report the number of iterations in MM 
and \ourmethod in Table~\ref{table:CCPvsGMM_cntIters}. The 
results for \ourmethod depend on the value of the progress 
coefficient $\eta$ which is set to match the experiments in the 
paper; $\sq{\eta=0.02}$ for the clustering experiment (Section 
\ref{sec:experiments:clustering}) and $\sq{\eta=0.10}$ for the 
scene recognition experiment 
(Section~\ref{sec:experiments:scene_classification}).

The overhead of the bound construction step depends on the
application. For example, in the scene recognition experiment, 
optimizing the bounds takes orders of magnitude more than 
sampling them (a couple of hours vs. a few seconds). In the 
clustering experiment, however, the optimization step is 
solved in closed form whereas sampling a bound involves 
performing a random walk on a large graph which can take a 
couple of minutes to run.

\begin{table}
\small
\centering
	\renewcommand{\tabcolsep}{4pt}
	\begin{tabular}{| c | c | c | c  | c | }
	\hline
	\multirow{2}{*}{\textbf{experiment}} & \multirow{2}{*}{\textbf{setup}} &
	\multirow{2}{*}{\textbf{MM}} & \multicolumn{2}{c|}{\textbf{\ourmethod}}
	\\ \cline{4-5}
	& & & \textbf{random} & \textbf{biased}
	\\ \hline
	scene & $\lambda=0.0$ & 145 & 107 & 87
	\\ \cline{2-5}
	recognition & $\lambda=1.0$ & 65 & 69 & 138
	\\ \hline \hline
	data & forgy & $35.76 \pm 7.8$ & \multicolumn{2}{c|}{$91.52 \pm 4.4$}
	\\ \cline{2-5}
	clustering & rand. part. & $114.98 \pm 12.9$ & \multicolumn{2}{c|}{$241.89 \pm 2.1$}
	\\ \cline{2-5}
	(GMM-200) & \kmeanspp & $32.92 \pm 5.8$ & \multicolumn{2}{c|}{$80.78 \pm 2.9$}
	\\ \hline
	 data & forgy & $37.18 \pm 12.1$ & \multicolumn{2}{c|}{$87.68 \pm 15.4$}
	\\ \cline{2-5}
	 clustering & rand. part. & $65.14 \pm 18.7$ & \multicolumn{2}{c|}{$138.64 \pm 5.9$}
	\\ \cline{2-5}
	 (Cloud) & \kmeanspp & $21.3 \pm 4.1$ & \multicolumn{2}{c|}{$44.12 \pm 10.7$}
	\\ \hline
	\end{tabular}
	\vspace{-0.7em}
\caption{Comparison of the number of iterations that MM and \ourmethod take to converge in the scene recognition and the data clustering experiment with different initializations. The numbers reported for the clustering experiment are the average and standard deviation over $50$ trials.}
\label{table:CCPvsGMM_cntIters}
\vspace{-1em}
\end{table}

\section{Conclusion}
We introduced Generalized Majorization-Minimization (\ourmethod), 
an iterative bound optimization framework that generalizes 
Majorization-Minimization (MM). Our key observation is that MM 
enforces an overly-restrictive touching constraint when constructing 
bounds, which is inflexible and can lead to sensitivity to initialization. 
By adopting a different measure of progress, \ourmethod relaxes this 
constraint, allowing more freedom in bound construction. Specifically, 
we propose deterministic and stochastic ways of selecting bounds 
from a set of valid ones. This generalized bound construction process 
tends to be less sensitive to initialization, and enjoys the ability to 
directly incorporate rich application-specific priors and constraints, 
without modifications to the objective function. In experiments with 
several latent variable models, \ourmethod algorithms are shown to 
significantly outperform their MM counterparts.

Future work includes applying \ourmethod to a wider range of 
problems and theoretical analysis, such as convergence rate. We 
also note that, although \ourmethod is more conservative than MM in 
moving towards nearby local minima, it still requires \emph{making 
progress} in every step. Another interesting research direction is to 
enable \ourmethod to occasionally pick bounds that do not make 
progress with respect to the solution of the previous bound, thereby 
making it possible to get out of local minima, while still maintaining 
the convergence guarantees of the method.


\bibliography{gmm_bibfile}

\section*{Appendix}
In this appendix we will provide proofs for the two theorems that we 
presented in the main paper. We also provide more visualization of 
the models trained with \ourmethod and compare them with CCCP 
and EM, which we had to omit from the main paper due to space 
limitations.

\appendix
\section{Proof of Convergence}

\begin{proof}[Proof of Theorem \ref{thrm:gmm_sol_converges}.]  
First, we observe that the following inequality follows from the bound construction assumptions:
\begin{align}
b_t(w_t) \leq b_t(w_{t-1}) \leq v_{t-1},
		\label{eq:easy_bound}
\end{align}
where  $v_t = b_t(w_t)-\eta d_t$. In particular, the first inequality holds because $w_t$ minimizes $b_t$ and the second inequality follows from \eqref{eq:valid_bounds_eta}.  Summing \eqref{eq:easy_bound}	over $t=1,...,T$ and substituting the definition of $v_t$ gives
\[
		\sum_{t=1}^T b_t(w_t) \leq \sum_{t=1}^T v_{t-1} = v_0 + \sum_{t=1}^{T-1} \big(b_t(w_t) - \eta d_t\big)
\]
which implies
\begin{equation}
	\eta \sum_{t=1}^T d_t \leq v_0 - b_T(w_T).
	\label{eq:touching_inequality}
\end{equation}
Recall that we set $v_0=F(w_0)$, and let $ F_*\in\mathbb{R}$ denote a finite global lower bound for $F$, and hence $b_T(w_T)\geq F_*$. The bound \eqref{eq:touching_inequality} then implies 
\[
\eta \sum_{t=1}^\infty d_t \leq F(w_0)-F_* < \infty,
\]
which gives $\lim_{t\to\infty } d_t =0. $

Next, recall that for every $m$-strongly convex function $f$, every $x,y$ in the domain of $f$, and every subgradient $g \in \partial f(x)$, we have
	\begin{align}
		f(y) \geq f(x) + g^T (y-x) + \frac{m}{2} || x - y ||^2 .
		\label{eq:functions_with_bounded_curvature}
	\end{align}
Substituting $f = b_t$, $x = w_t$, and $y = w_{t-1}$ in (\ref{eq:functions_with_bounded_curvature}), and noting that the zero vector is a subgradient of $b_t$ at $w_t$ (because $w_t$ is a minimizer of $b_t$), we obtain
\begin{align}
	 || w_t - w_{t-1} ||^2  &\leq \frac{2}{m} \left( b_t(w_{t-1}) - b_t(w_t) \right)\notag\\
	 & \leq  \frac{2}{m} \left( b_{t-1}(w_{t-1}) - b_t(w_t) \right),
	\label{eq:bounds_with_bounded_curvature}
\end{align}
where (\ref{eq:valid_bounds_eta}) is used in the second inequality. Summing \eqref{eq:bounds_with_bounded_curvature} over $t=2,..,T$, we obtain 
\begin{align}
	\sum_{t=1}^T || w_t - w_{t-1} ||^2 & \leq b_1(w_1)-b_T(w_T)\notag\\
	& \leq F(w_1)-F_*,
	\label{eq:gmm_converges}
\end{align}
which implies 
\begin{equation}
	\lim_{t\to\infty}\|w_t-w_{t-1}\|=0
	\label{eq:w_limit}
\end{equation}
On the other hand, since $F(w_t)\leq b_t(w_t) \leq F(w_0)$ by \eqref{eq:GMMconstraint}, the sequence $\{w_t\}_t$ lies in the sublevel set $\{w\in\mathbb{R}^n | F(w) \leq F(w_0)\}$, which is assumed to be a compact set. To show that a sequence that is contained in a compact set converges, one needs to prove that all its converging subsequences have the same limit. For $\{w_t\}_t$, this follows from \eqref{eq:w_limit}, and therefore $\{w_t\}_t$ converges to a limit $w^\dagger$.
\end{proof}

\begin{figure}
\centering
	\includegraphics[width=0.40\textwidth]{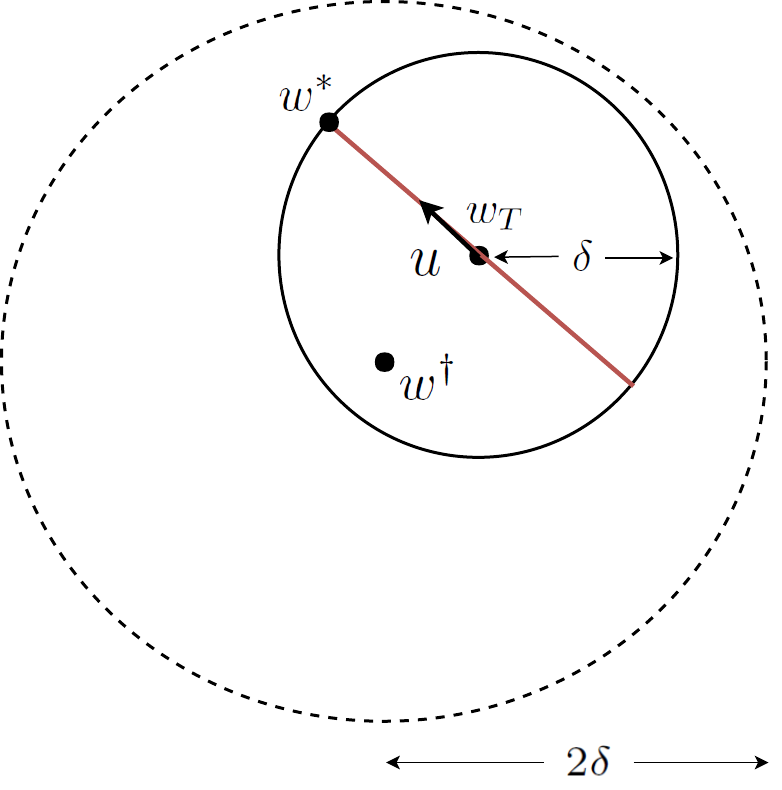}
	\caption{An illustration of quantities defined in the proof of Theorem \ref{thrm:GMM-converges-to-local-min}}
\label{fig:proof}
\end{figure}

\begin{proof}[Proof of Theorem \ref{thrm:GMM-converges-to-local-min}.] 
We prove this theorem by contradiction. Suppose $\nabla F(w^\dagger)\neq 0$. This implies that there exists a unit vector $u\in\mathbb{R}^d$ such that the directional derivative of $F$ along $u$ is positive at $w^\dagger$, \ie $\nabla F(w^\dagger)\cdot u>2c$ for some $c>0.$ Since $F$ is continuously differentiable, $\nabla F\cdot u$ is continuous at $w^\dagger$, and hence 
\begin{equation}
\nabla F(w)\cdot u>c,\quad\quad \forall w\in B_{2\delta}(w^\dagger),
\label{eq:directional_derivative}
\end{equation}
for all small enough $\delta>0$, where $B_{r}(x)$ denotes an open ball around $x$ with radius $r$. We fix a $\delta>0$ that satisfies \eqref{eq:directional_derivative}, as well as the bound
\begin{equation}
 \delta<\frac{2c}{M}.
 \label{eq:delta_condition}
\end{equation}
 We also fix an $\epsilon>0$ that satisfies
\begin{equation}
 \epsilon< c\delta-\frac{M}{2}\delta^2,
 \label{eq:epsilon}
\end{equation}
which is possible because of \eqref{eq:delta_condition}. The reason for this will be clear shortly.

Now recall by Theorem \ref{thrm:gmm_sol_converges} that $w_t\to w^\dagger$ and $d_t\to 0$, as $t\to\infty$, so we can pick $T>0$ large enough such that 
\begin{equation}
|w_T-w^\dagger|<\delta
\end{equation}
and 
\begin{equation}
d_T = b(w_T)-F(w_t) <\epsilon
\label{eq:bound_dT}
\end{equation}
Now define the function $g$ to be the restriction of $F$ on a line parallel to $u$ that passes through $w_T$ (see Figure~\ref{fig:proof}), that is 
\[
g(z) = F(w_T+zu),\quad\quad z\in\mathbb{R}.
\]
It is easy to see that $g$ is continuously differentiable with 
\[
g'(z) = \nabla  F (w_T+zu) \cdot u.
\]
In particular, the bound \eqref{eq:directional_derivative} implies 
\begin{equation}
g'(z) >c,\quad\quad z\in(0,\delta).
\label{eq:gprime_bound}
\end{equation}
This is because for every $z\in(0,\delta)$,
\[w_T+zu\in B_\delta (w_T) \subset B_{2\delta}(w^\dagger).\]

An application of Taylor Expansion Theorem of order $n=0$ on $g$ around $z=0$ shows that there exits a $z_*\in(0,\delta)$ such that
\begin{align*}
g(\delta) &= g(0)+ g'(z_*) \delta > g(0)+c \delta,
\end{align*}
where we used $g'(z_*)>c$ by \eqref{eq:gprime_bound}. Substituting definitions of $g(0)$ and $g'(0)$ in the display above, we obtain the bound
\begin{equation}
F(w_*)> F(w_T)+c\delta, \quad w_* = w_T + \delta u. 
\label{eq:F_lowerbound}
\end{equation} 

On the other hand, since $b_T$ is a smooth function with its minimum at $w_T$ and its Hessian $\nabla^2 b_T$ bounded by $MI$, second order Taylor expansion of $b_T$ around $w_T$ gives
\[
b_T(w) \leq b_T(w_T)+\nabla b_T(w_T)\cdot(w-w_T) + \frac{M}{2}\|w-w_T\|^2,
\]
and in particular, for $w=w_*=w_T+\delta u$,
\begin{equation}
b_T(w_*) \leq b_T(w_T) + \frac{M}{2}\delta^2.
\label{eq:bound_bT}
\end{equation}

Combining the bounds \eqref{eq:bound_dT}-\eqref{eq:bound_bT} and the choice \eqref{eq:epsilon} of $\epsilon$, we have
\begin{align*}
b_T(w_*)-F(w_*) & \leq [b_T(w_T)-F(w_T)]  + \frac{M}{2}\delta^2 - c\delta\\
& \leq  \epsilon + \frac{M}{2}\delta^2 - c\delta\\
& <0,
\end{align*}
which contradicts the fact that $b_T$ is an upper bound for $F$. This completes the proof.
\end{proof}

\section{\kmeans Clustering}

\begin{figure*}
    \centering
    \begin{subfigure}[b]{0.3\textwidth}
        \centering
        \includegraphics[width=\textwidth]{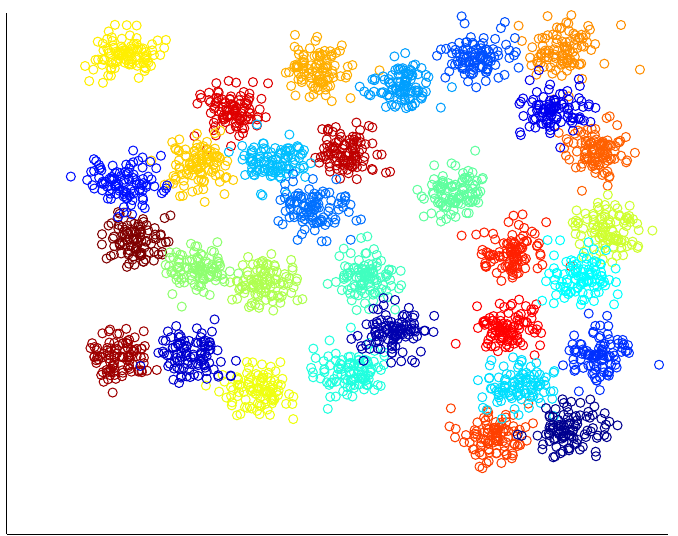}
        \caption{ground truth}
    \end{subfigure}
    \hfill
    \begin{subfigure}[b]{0.3\textwidth}
        \centering
        \includegraphics[width=\textwidth]{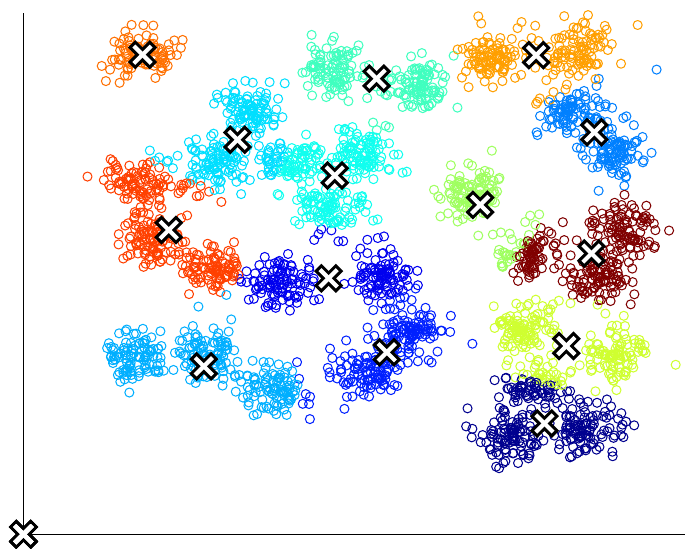}
        \vspace{-5.9mm}
        \caption{\kmeans}
    \end{subfigure}
    \hfill
    \begin{subfigure}[b]{0.3\textwidth}
        \centering
        \includegraphics[width=\textwidth]{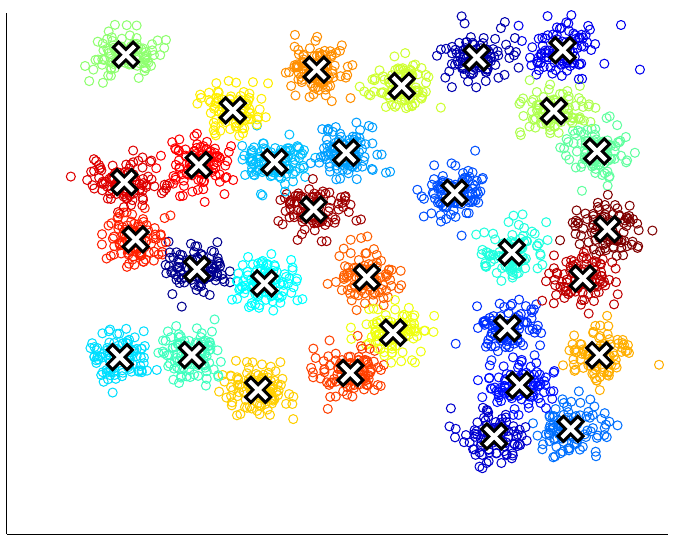}
        \caption{generalized-MM}
    \end{subfigure}
    \caption{Visualization of clustering solutions on the D31 dataset~\cite{kmeans:D31} 
from identical initializations. Random partition initialization scheme is used. (a) color-coded ground-truth 
clusters. (b) solution of \kmeans. (c) solution of \ourmethod. The white crosses indicate location of the cluster 
centers. Color codes match up to a permutation.}
    \label{fig:kmeans_visualization}
\end{figure*}

Figure~\ref{fig:kmeans_visualization} visualizes the result of 
\kmeans and \ourmethod (with random bounds) on the D-31 dataset 
\cite{kmeans:D31}, from the same initialization. \ourmethod finds a 
near perfect solution,  while in standard \kmeans, many clusters get 
merged incorrectly or die off. Dead clusters are those which do not 
get any points assigned to them. The update rule (M-step of \kmeans 
algorithm) collapses the dead clusters on to the origin.

\section{\lssvm for Mammal Image Classification}
\begin{figure*}
\centering
\includegraphics[height=6.8em]{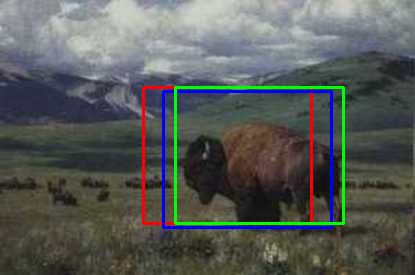}
\includegraphics[height=6.8em]{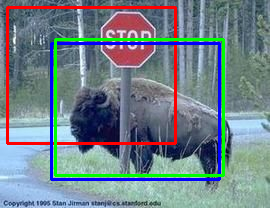}
\includegraphics[height=6.8em]{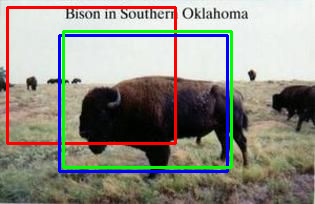}
\includegraphics[height=6.8em]{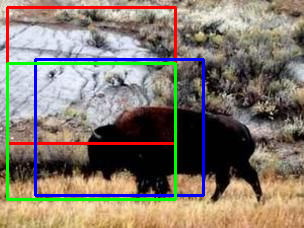}

\includegraphics[height=6.8em]{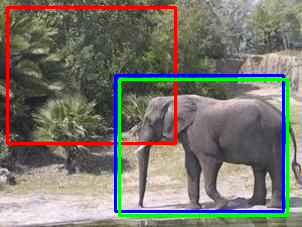}
\includegraphics[height=6.8em]{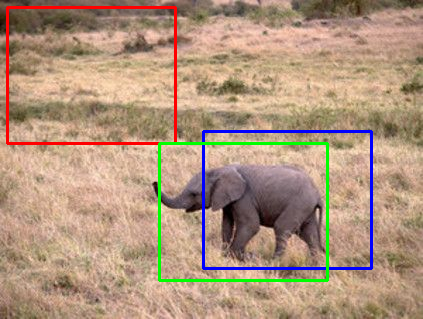}
\includegraphics[height=6.8em]{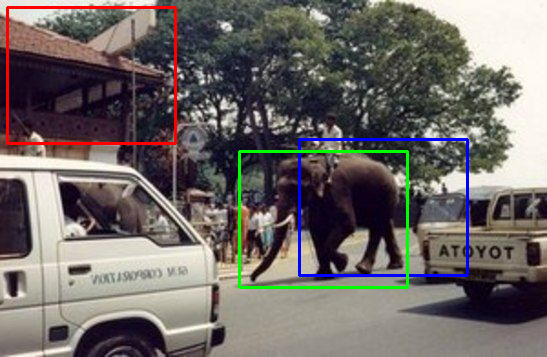}
\includegraphics[height=6.8em]{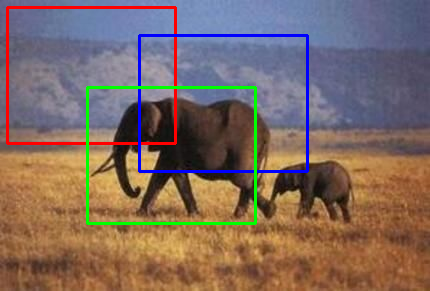}

\includegraphics[height=6.8em]{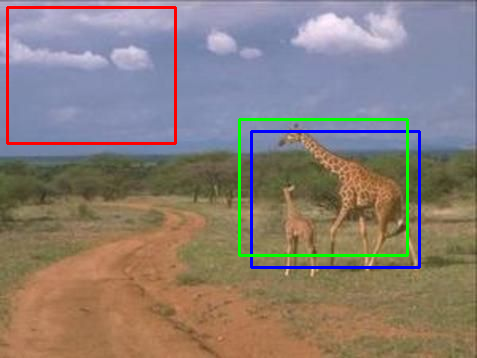}
\includegraphics[height=6.8em]{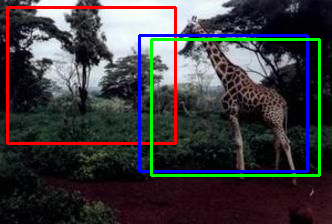}
\includegraphics[height=6.8em]{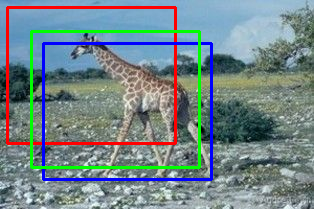}
\includegraphics[height=6.8em]{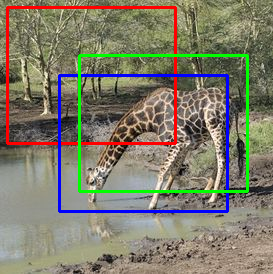}

\includegraphics[height=6.8em]{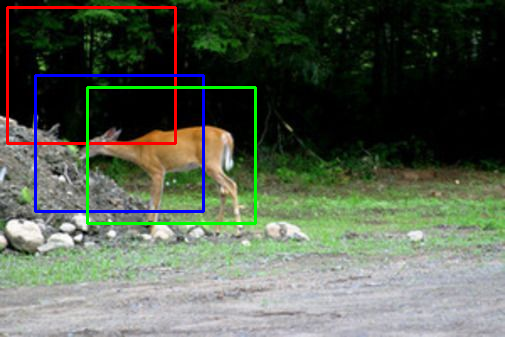}
\includegraphics[height=6.8em]{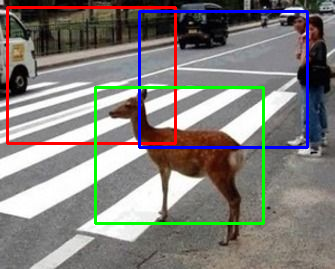}
\includegraphics[height=6.8em]{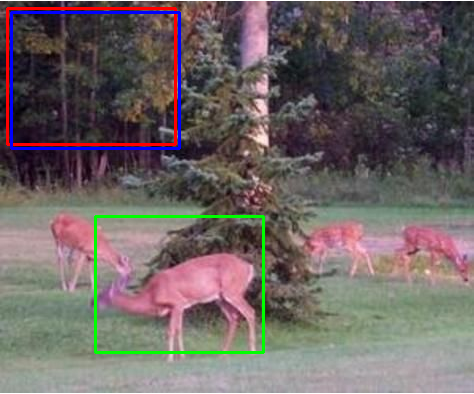}
\includegraphics[height=6.8em]{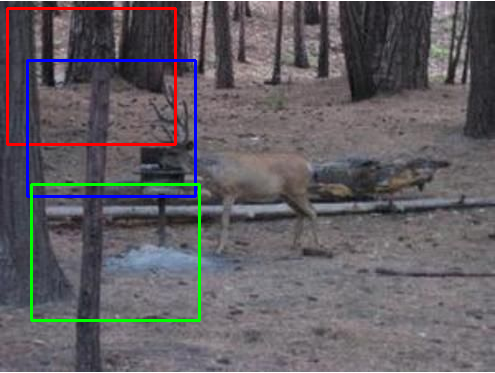}

\includegraphics[height=6.8em]{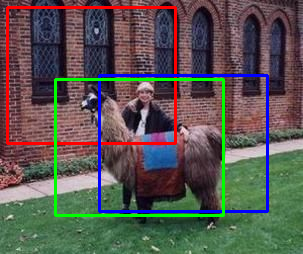}
\includegraphics[height=6.8em]{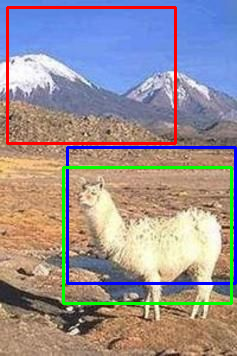}
\includegraphics[height=6.8em]{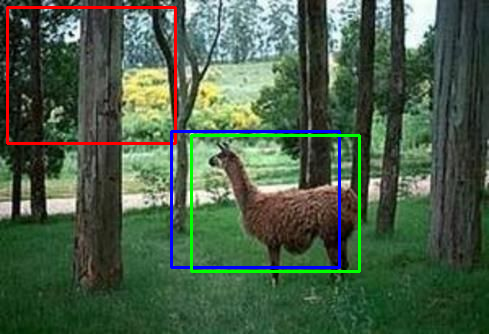}
\includegraphics[height=6.8em]{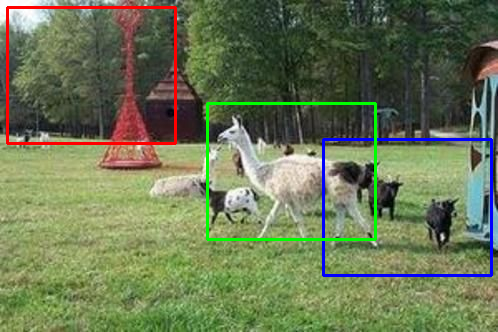}

\includegraphics[height=6.8em]{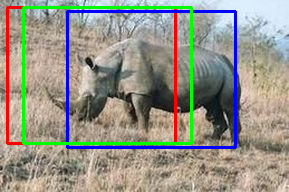}
\includegraphics[height=6.8em]{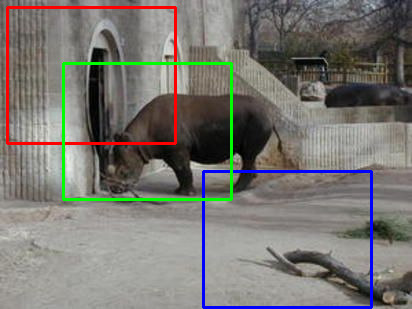}
\includegraphics[height=6.8em]{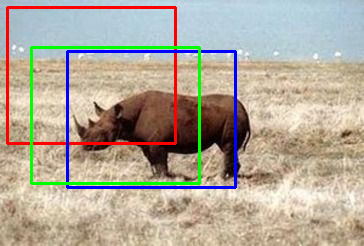}
\caption{Example training images from the mammals dataset, shown 
	with final imputed latent object locations by three algorithms: CCCP 
	(red), G-MM random (blue), G-MM biased (green). Initialization: 
	\emph{top-left}.}
\label{fig:mammals}
\end{figure*}

We provide additional experimental results on the mammals dataset. 
Figure~\ref{fig:mammals} shows example training images and the 
final imputed latent object locations by three algorithms: CCCP (red), 
G-MM random (blue), and G-MM biased (green). The initialization is 
\emph{top-left}.

In most cases CCCP fails to update the latent locations given by 
initialization. The two G-MM variants, however, are able to update 
them significantly and often localize the objects in training images 
correctly. This is achieved \emph{only} with image-level object 
category annotations, and with a very bad (even adversarial) 
initialization.

\bibliographystyle{icml2019}

\end{document}